\documentclass[runningheads]{llncs}
\usepackage{graphicx}

\usepackage{tikz}
\usepackage{comment}
\usepackage{amsmath,amssymb} 
\usepackage{color}
\usepackage{multirow}
\usepackage[linesnumbered,ruled,vlined]{algorithm2e}
\usepackage{booktabs}
\usepackage{url}
\usepackage[accsupp]{axessibility}  

\begin{document}
	\pagestyle{headings}
	\mainmatter
	\def\ECCVSubNumber{7417}  
	
	\title{3D Random Occlusion and Multi-Layer Projection for Deep Multi-Camera Pedestrian Localization} 

	\titlerunning{3DROM}
	%
	\author{Rui Qiu\inst{1,2} \and
		Ming Xu\inst{1,2,*} \and
		Yuyao Yan\inst{1} \and
		Jeremy S. Smith\inst{2} \and
		Xi Yang\inst{1}}
	\authorrunning{R. Qiu, M. Xu et al.}
	%
	\institute{School of Advanced Technology, Xi'an Jiaotong-Liverpool University, Suzhou, 215123,
		China \\
		\email{\{ming.xu, xi.yang01\}@xjtlu.edu.cn} \and
		Department of Electrical Engineering and Electronics, University of Liverpool, Liverpool, L69 3BX, UK \\
		\email{\{rui.qiu, j.s.smith\}@liverpool.ac.uk}}
	
	\maketitle
	
	\renewcommand{\thefootnote}{}
	\footnotetext{* M. Xu is the corresponding author.}
	
	\begin{abstract}
		Although deep-learning based methods for monocular pedestrian detection have made great progress, they are still vulnerable to heavy occlusions. Using multi-view information fusion is a potential solution but has limited applications, due to the lack of annotated training samples in existing multi-view datasets, which increases the risk of overfitting. To address this problem, a data augmentation method is proposed to randomly generate 3D cylinder occlusions, on the ground plane, which are of the average size of pedestrians and projected to multiple views, to relieve the impact of overfitting in the training. Moreover, the feature map of each view is projected to multiple parallel planes at different heights, by using homographies, which allows the CNNs to fully utilize the features across the height of each pedestrian to infer the locations of pedestrians on the ground plane. The proposed 3DROM method has a greatly improved performance in comparison with the state-of-the-art deep-learning based methods for multi-view pedestrian detection. Code is available at \url{https://github.com/xjtlu-cvlab/3DROM}.
		
	\end{abstract}

	\section{Introduction}
	
	
	Pedestrian detection plays an important role in the fields of tracking, person re-identification and crowd counting.
	In recent years, deep-learning based object detection methods have made significant progress in pedestrian detection.
	However, these deep monocular methods are not robust enough to detect heavily occluded pedestrians or localise partially occluded pedestrians on the ground. The solution to this problem lies in multi-view pedestrian detection. Compared with single-view pedestrian detection, multi-view methods can detect heavily occluded pedestrians more effectively and accurately \cite{fleuret2007multicamera}.
	
	
	Deep-learning based multi-camera detection methods need to be trained on sufficient annotated samples to achieve the desired performance. However, the limited ground truth data available in existing multi-view video datasets makes it difficult for the network to achieve the best performance in training, which limits deep learning methods from being widely used in multi-view pedestrian detection. The reason behind this is that the annotation of a multi-view pedestrian dataset is a tedious and time-consuming process. For example, with the help of an annotation tool specifically designed for multi-view datasets, it took a trained annotator an average of 10 minutes to annotate one frame with 7 views for the WILDTRACK dataset \cite{WILDTRACK_dataset}\cite{chavdarova2018wildtrack}. On the other hand, although monocular data augmentation methods, such as flipping, random cropping and Random Erasing~\cite{zhong2020random}, can relieve overfitting and improve the robustness of the networks to occlusion, these methods violate the homographic constraint among multiple views and cannot be used for multi-view pedestrian detection methods.
	
	\begin{figure}[t]
		\centering
		\includegraphics[width=4.87in]{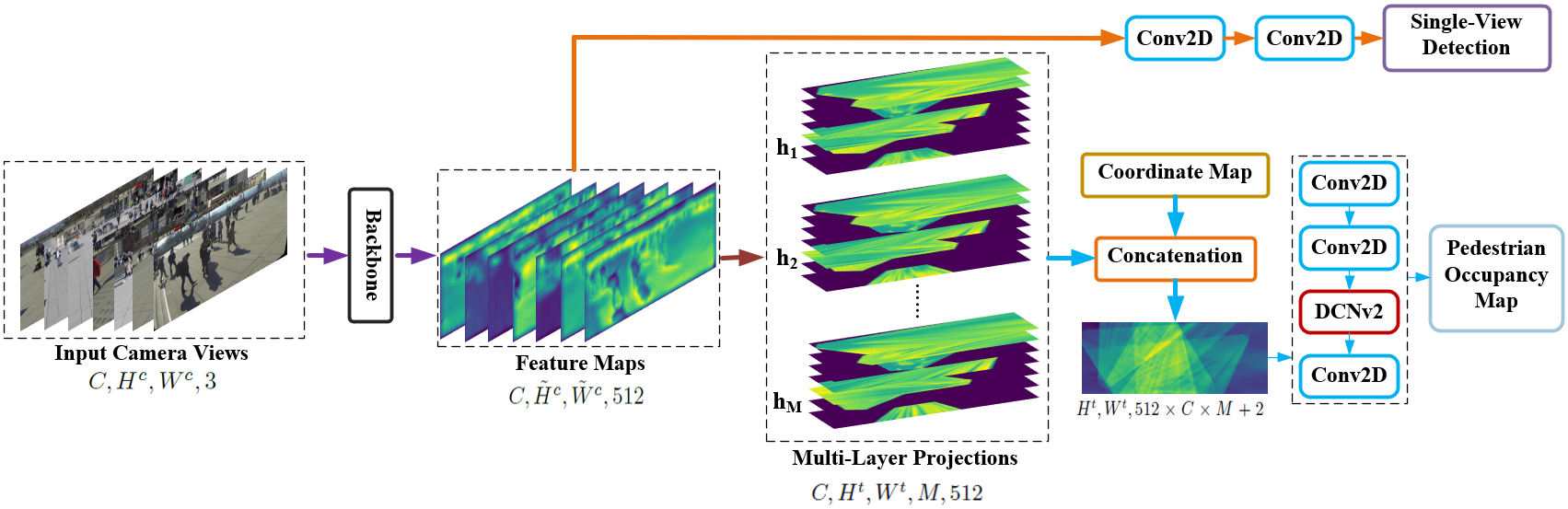}
		\caption{The structure of 3DROM. $h_1$, $h_2$, \dots, $h_M$ represent the multi-layer projections at different heights.}
		\label{f1}
	\end{figure}
	In this paper, on the basis of the MVDet framework \cite{hou2020multiview}, a data augmentation method is proposed to address this problem, in which occlusion boxes are randomly but consistently added to multiple camera views in the training. In this method, the ground-plane area of interest (AOI) is discretized into a grid of locations; 3D cylinders, of the average size of pedestrians, are placed at randomly selected locations on the ground plane and projected into each of the multiple camera views as filled rectangles. It reduces the risk of overfitting in the training and improves the robustness of pedestrian detection with heavy occlusions. In addition, the feature maps are projected to multiple planes parallel to the ground plane and at different heights.
	The multi-layer projection allows the different features (feet, torso and head) of each pedestrian to be projected to the same location in the top view but at different heights. This allows the features across the height of that pedestrian to be fully utilised in comparison with the ground-plane feature projection in MVDet. This proposed algorithm is referred to as 3DROM. A schematic diagram of the system architecture is shown in Fig.~\ref{f1}.
	
	The contributions of this paper are twofold:
	(1) A data augmentation method is proposed for deep multi-view pedestrian detection, in which 3D random occlusions are generated and back-projected to multiple camera views. It can be used to prevent overfitting and improve the detection performance with a limited number of multi-view training samples.
	To the best of our knowledge, this method is used for deep multi-view pedestrian detection for the first time.
	(2) A multi-layer projection method for the single-view feature maps is used to fully utilize the pedestrians' features across a range of heights. The locations of pedestrians can be inferred from the multi-height features, rather than only the ground-plane features, of the pedestrians.
	
	
	\section{Related Work}
	\subsection{Multi-view Pedestrian Detection}
	A recent survey on multi-view pedestrian detection can be found in \cite{Qiu_book2022}. The state-of-the-art methods in this field can be categorised into top-down approach and bottom-up approach. The top-down approach divides the ground plane into a grid. Each location in this grid is thought of as the location of a potential pedestrian and is back-projected to individual views for finding the optimal match between foregrounds and a generative model. The bottom-up approach projects the foregrounds from the individual views to a reference view and analyses the overlaid foreground projections to determine the locations of pedestrians.

	\textbf{Top-Down Approach}
	Fleuret et al.~\cite{fleuret2007multicamera} estimated a probabilistic occupancy map 
	through a generative model that represents each pedestrian as a filled rectangle of the average size of pedestrians. The occupancy probability was updated iteratively for finding the locations of the rectangles which cover more foreground pixels in all the views.
	On the basis of this point of view, Alahi et al.~\cite{alahi2011sparsity} formulated multi-view pedestrian detection as a linear inverse problem; Peng et al.~\cite{peng2015robust} modelled pedestrians and their occlusion relationships by using a multi-view Bayesian network; Yan et al.~\cite{yan2021multicamera} used a non-iterative logic minimization method to reduce false-positive detections.
	Chavdarova and Fleuret \cite{chavdarova2017deep} proposed an end-to-end multi-view pedestrian detection network. They back-projected each ground-plane location to individual views and created a rectangle box at the corresponding positions. A CNN was used to extract features within these rectangles and infer the locations of pedestrians by using Multi-Layer Perception. 
	Baqu{\'e} et al.~\cite{baque2017deep} proposed a method which combines CNNs and a Conditional Random Field. The CNN in the discriminative model extracts pedestrian features from individual views and uses Gaussian Mixture networks to classify the body parts as pedestrian features. Meanwhile, a generative model is used to model the occlusion relationships among pedestrians. The locations where the discriminative model fits the generative model well are thought of as the locations of pedestrians.
	
	\textbf{Bottom-Up Approach}
	Khan and Shah \cite{KhanS06}\cite{khan2008tracking} projected the foreground likelihood maps of individual views to a reference view using multi-plane homographies. Areas with heavily overlaid foregrounds are thought of as the locations of pedestrians. However, the foreground projections of different pedestrians may overlap, which leads to false positive detections.
	Eshel and Moses \cite{eshel2010tracking} projected the individual views to the head plane and detected pedestrians at the locations where the intensities projected from different views are pixelwise correlated.
	Ge and Collins \cite{ge2010crowd} modelled each pedestrian as a cylinder and used Gibbs sampling to find the locations of pedestrians.
	Utasi and Benedek \cite{utasi2012bayesian} also used cylinders in the 3D space to model the foreground silhouettes, which was enhanced by pixel-level leg and head features, and determined the pedestrians' locations by using a 3D Bayesian Marked Point Process model.
	Xu et al.~\cite{xu2016multi} detected pedestrians in individual views using Faster RCNN~\cite{ren2016faster} and projected the foot points of the bounding boxes of the pedestrians to the ground plane. They clustered the projected foot points to determine the locations of pedestrians in the top view. 
	Hou et al. proposed MVDet \cite{hou2020multiview}, an anchor-free end-to-end pedestrian detection network. This system uses ResNet18 \cite{he2016deep} as the backbone to extract feature maps from individual views. The feature maps from multiple views are projected to the ground plane and concatenated there. Then a ground-plane classifier predicts the locations of pedestrians. This feature projection method is similar to that proposed by Zhang and Chan~\cite{zhang2019wide}\cite{zhang2021cross} for multi-camera crowd counting.
	On the basis of the MVDet framework, Song et al.~\cite{song2021stacked} proposed the SHOT algorithm which projects the feature map of each individual view to multiple parallel planes. The multi-plane feature maps projected from the same view were weighted and summed into one feature map. Such feature maps from the multiple views are concatenated to predict a pedestrian occupancy map on the ground. When the multi-height feature maps were summed into a single feature map, it causes an information loss; whilst such multi-height feature maps are concatenated with no information loss in the 3DROM algorithm, which leads to an improved performance.
	

	\subsection{Data Augmentation}
	In deep-learning based methods, data augmentation methods are widely used to increase the number of training samples and improve the robustness by applying various transformations to existing samples \cite{he2016deep}\cite{simonyan2014very}\cite{krizhevsky2012imagenet}. One of these methods is to directly apply an image processing operation, such as flipping, folding, rotating, adding noise and Random Erasing, to existing samples. Random Erasing ~\cite{zhong2020random} overwrites each pixel in a randomly selected region of an image with a random colour. This method can be applied to the training of deep-learning based algorithms for image classification, person re-identification and object detection tasks. It improves the robustness of an algorithm to occlusion and reduces the risk of overfitting the samples in the training. In addition, Wang et al.~\cite{wang2017fast} proposed a method for generating samples with occlusion and deformation using adversarial networks. These generated samples can improve the accuracy and robustness of Faster R-CNN in the detection of deformed or occluded objects. However, both methods are currently used in monocular detection only and cannot work well for deep end-to-end multi-view pedestrian detection without considering the geometrical relationship among multiple views.

	
	\begin{figure}[t]
		\centering
		\includegraphics[width=4.5in]{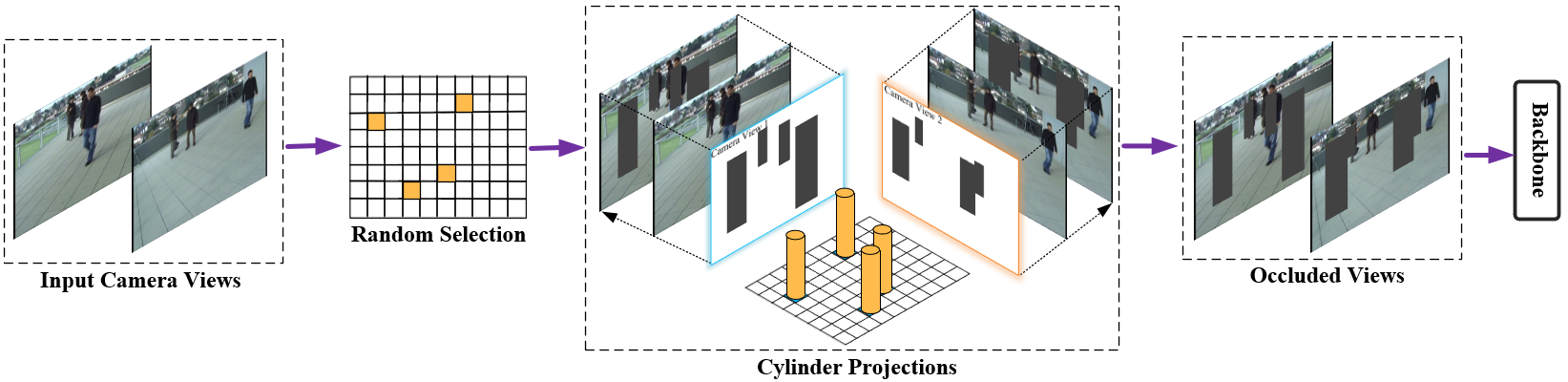}
		\caption{A schematic diagram of the 3D Random Occlusion method.}
		\label{f2}
	\end{figure}

	\section{Methodology}
	
	The motivation of our work is to address the performance improvement on deep multi-view pedestrian detection networks with a limited number of training samples.
	Robust pedestrian detection requires efficient network training with limited samples and an effective fusion method for multi-view features. We focus on reducing the risk of overfitting during the training and improving the utilization of the feature maps across multiple views to improve the detection performance in the MVDet framework.
	
	\subsection{Notations and Homography Estimation}
	Let $C$ be the number of the cameras in a multi-view pedestrian dataset. The size of input image $I_c$, from camera view $c$ $(c\in [1, C])$, is $H^c \times W^c$. $(u^c,v^c)$ represents an image coordinate in view $c$. The size of the feature map $F_c$, extracted from camera view $c$, is $\widetilde{H}^c \times \widetilde{W}^c$. $H^t \times W^t$ is the size of the top view image. Assume the area of interest (AOI) on the ground plane is discretized into a grid of $G$ locations. Let $\textbf{X}_i$ be the coordinate of the $i$-th location $(i \in [1, G])$. Let $S$ denote the set of the index numbers for the grid locations that have been selected to place the 3D occlusions.
	
	Planar homography is the relationship between a pair of captured images of the same plane.	
	Let $\textbf{u}$ and $\textbf{X}$ be the homogeneous image coordinates of the same point on a plane in  camera view $c$ and the top view. They are associated by the homography matrix $\textbf{H}^{c,t}$ for that plane as follow:
	\begin{equation}
		\textbf{X} \cong \textbf{H}^{c,t}\textbf{u}.
	\end{equation}
	
	A $3 \times 4$ projection matrix can be calculated by using the intrinsic and extrinsic parameters of  camera $c$: $\textbf{M} = [\textbf{m}_1,\textbf{m}_2,\textbf{m}_3,\textbf{m}_4]$. 	
	The homography matrix, from the top view $t$ to camera view $c$, for the ground plane is:
	\begin{equation}
		\textbf{H}^{t,c}_{0} = (\textbf{H}^{c,t}_{0})^{-1} = [\textbf{m}_1,\textbf{m}_2,\textbf{m}_4].
	\end{equation}
	
	The homography matrix, from the top view $t$ to camera view $c$, for the plane parallel to the ground plane and at a height of $h$ can be written as:
	\begin{equation}
		\textbf{H}^{t,c}_{h} = [\textbf{m}_1,\textbf{m}_2,h\textbf{m}_3 + \textbf{m}_4] = \textbf{H}^{t,c}_{0} +  [\textbf{0}\mid h\textbf{m}_3],
	\end{equation}
	where $[\textbf{0}]$ is a $3 \times 2$ zero matrix.

	\subsection{3D Random Occlusion}		
	Compared with single-view detection, multi-view pedestrian detection requires the use of geometric constraints to establish the correspondence among multiple views. Monocular data augmentation methods, such as flipping,  cropping, rotation and Random Erasing, may affect the performance of multi-view detection algorithms, since they violate the homography constraint. Therefore, Algorithm~\ref{code:alg1} was developed as a 3D data augmentation method for the training of multi-view pedestrian detection algorithms.
	
	The 3D Random Occlusion algorithm is based on the input camera views in the training, as shown in Algorithm~\ref{code:alg1}. The process of 3D Random Occlusion is illustrated in Fig.~\ref{f2}. The ground plane is discretized into a grid of locations. The $i$-th location $(i \in [1, G])$ in the top view is associated with its corresponding location $(u^c_i,v^c_i)$ in camera view $c$ $(c\in [1, C])$ through the ground-plane homography $\textbf{H}^{t,c}_0$. A 3D cylinder placed at the $i$th location on the ground plane is back-projected to a filled rectangle $r_i^c$ sitting at location $(u_i^c, v_i^c)$ in camera view $c$. The rectangle is designed to have the average height $H_i^c$ and width $W_i^c$ of the pedestrians standing at the $i$th location. $H_i^c$ is calculated as follows: the $i$-th location in the top view is projected back to camera view $c$ using the homographies, $\textbf{H}^{t,c}_0$ and $\textbf{H}^{t,c}_{h_a}$, for the planes at the heights of 0 cm and 180 cm; The vertical distance between the two projected points in view $c$ is $H_i^c$; the average width $W_i^c=\alpha H_i^c$, where $\alpha$ is a constant ratio.
	
	The inputs of Algorithm~\ref{code:alg1} are the images $I=\{I_1, I_2,\cdots, I_c\}$ from multiple camera views, the number of occlusions $n$ per frame and the occlusion probability $p$ of each frame to be selected to add 3D random occlusions. The $n$ locations in the top view are selected to generate filled rectangles at the corresponding locations in all the views. To ensure that the occlusions are not too close to each other, the ground distance between each selected location and other cylinder occlusions must be greater than a threshold $d=1$ meter. The selected locations are projected to all the views, by using homographies $\textbf{H}^{t,c}_0$ and $\textbf{H}^{t,c}_{h_a}$, to generate the filled rectangles with a constant pixel value $\Omega$.

	\subsection{The Multi-Layer Projection of Feature Maps}
	Within the MVDet framework, the feature map $F_c$ of view $c$ is projected to the ground plane by using a homography transformation. The feature map in the output of the backbone network does not have the same size as the input image $I_c$ of view $c$. However, it is resized to the same size afterwards. Therefore, the projected feature map $F_h^{c,t}$ from view $c$ to the top view can be written as:
	\begin{equation}
		F_h^{c,t} = \textbf{H}^{c,t}_{h}(F_c),
	\end{equation}
	for a plane parallel to the ground and at a height of $h$.

	The feature map on the ground plane is compromised when pedestrians' feet are occluded or their feet are off the ground. This may affect the model to infer the locations of the pedestrians. In \cite{khan2008tracking}, foreground likelihood maps are projected to multiple planes parallel to the ground plane and at different heights, which can significantly reduce detection errors. The foreground likelihood map, which indicates how likely a pixel in an image belongs to foregrounds, is similar to the feature map in MVDet. We assume that each pedestrian occupies a specific location in the top view. The multi-layer projection of the feature maps of multiple views can provide the comprehensive feature information for any pedestrian standing at that location. Compared with the ground-plane projection used in MVDet, the top view CNN is able to infer the locations of pedestrians from a wider range of features.
	\begin{center}
		\begin{minipage}[t]{10cm}
			\begin{algorithm}[H]
				\caption{3D Random Occlusion at one frame}  \label{code:alg1}
				\SetKwInOut{Input}{Input}
				\SetKwInOut{Output}{Output}
				\DontPrintSemicolon
				\Input{Input image $I=\{ I_1, I_2,\cdots, I_c \}$; \;
					The number of occlusions $n$ per frame; \;
					Occlusion probability $p$; \;
				}
				\Output{Occluded image $I^*=\{ I^*_1, I^*_2,\cdots, I^*_c \}$.}
				$S = \phi $; \;
				$I^* = I$; \;
				$p_1 =$ Rand(0, 1); \;
				\If{$p_1 > p$}
				{
					\Return $I^*$.
				}
				\Else
				{ $i = 0$; \;
					\While{$i < n$}
					{
						$k =$ Rand(1, $G$); \;
						\If{$\forall l\in S, \Vert \textbf{X}_k - \textbf{X}_l\Vert_2  < d$}
						{
							\textbf{goto} 9; \;
						}
						\Else
						{
							\For{camera view $c = 1$ to $C$}
							{
								$(\textbf{H}^{t,c}_{0}\textbf{X}_k,\textbf{H}^{t,c}_{h_a}\textbf{X}_k)\Rightarrow (u^c_k,v^c_k,H^c_k,W^c_k)$ ; \;
								\For{$u=u^c_k-W_k^c/2$ to $u_k^c+W_k^c/2$}
								{
									\For{$v=v^c_k$ to $v^c_k+H_k^c$}
									{
										$I^*_c(u, v) = \Omega$;\;
									}
								}
							}
							$S = S \cup \{k\}$;\;
							$i = i + 1$
						}
					}
					\Return $I^*$; \;}
			\end{algorithm}
		\end{minipage}
	\end{center}
	
\begin{figure}[t]
		\begin{center}
			\begin{tabular}{cc}
				\includegraphics[width=0.35\textwidth]{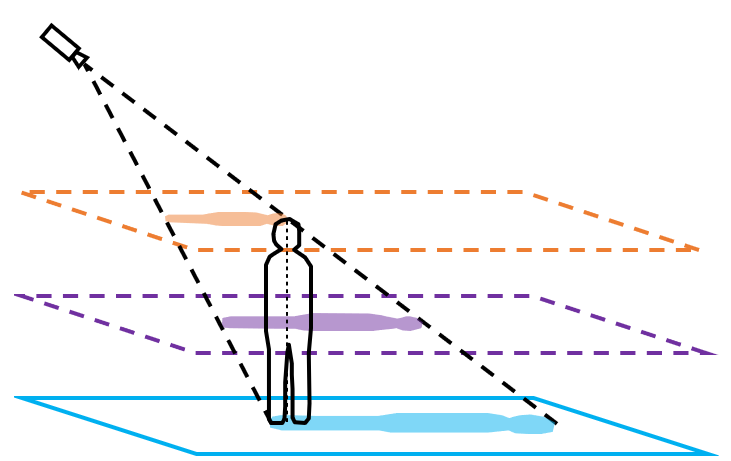}&
				\includegraphics[width=0.35\textwidth]{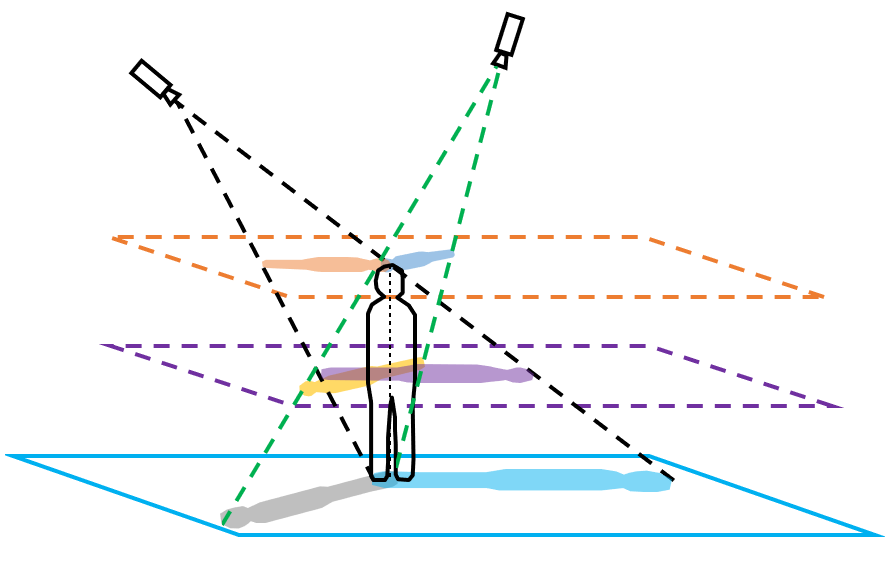}\\
				(a) & (b)\\
				\includegraphics[width=0.35\textwidth]{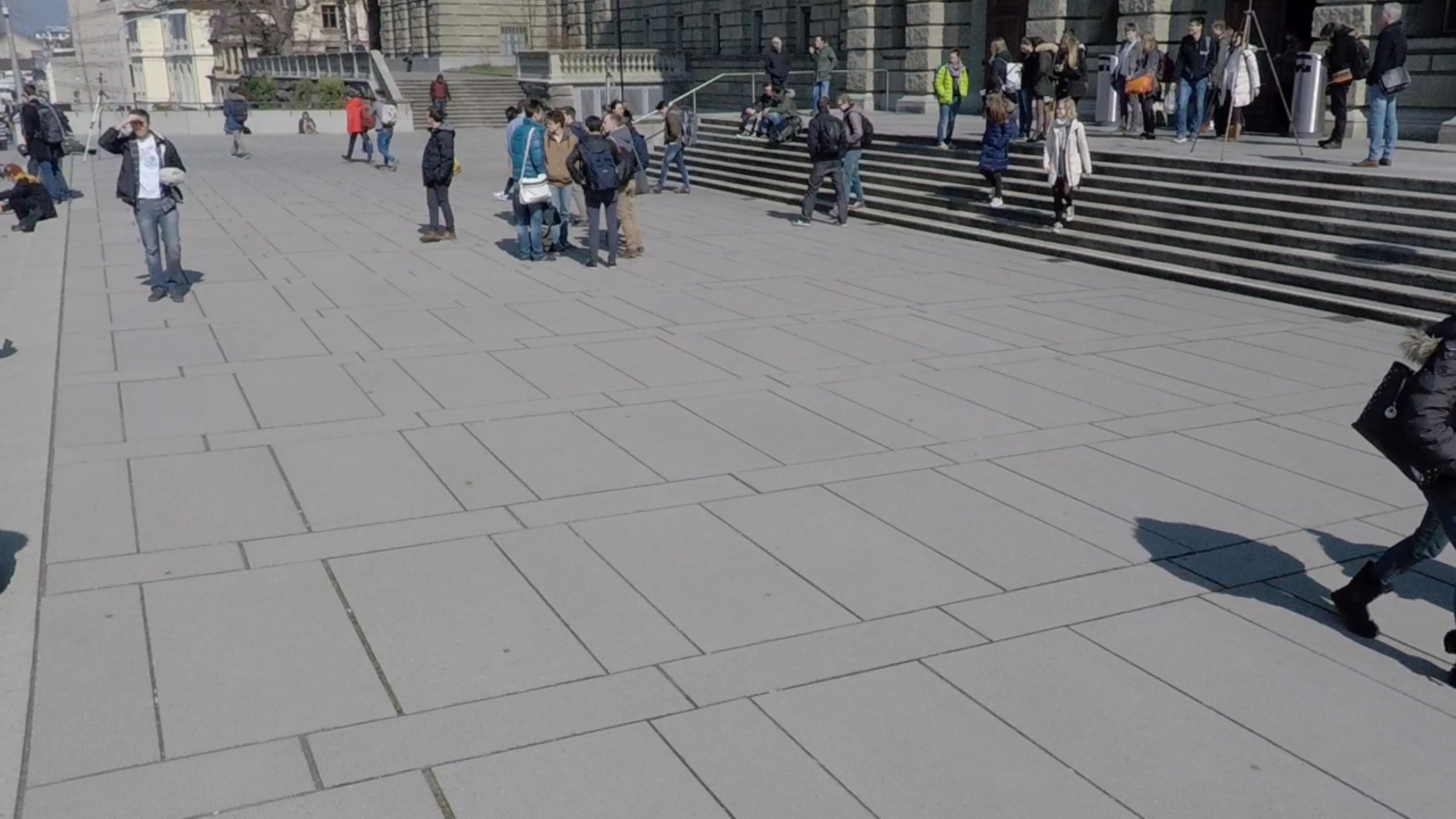}&
				\includegraphics[width=0.35\textwidth]{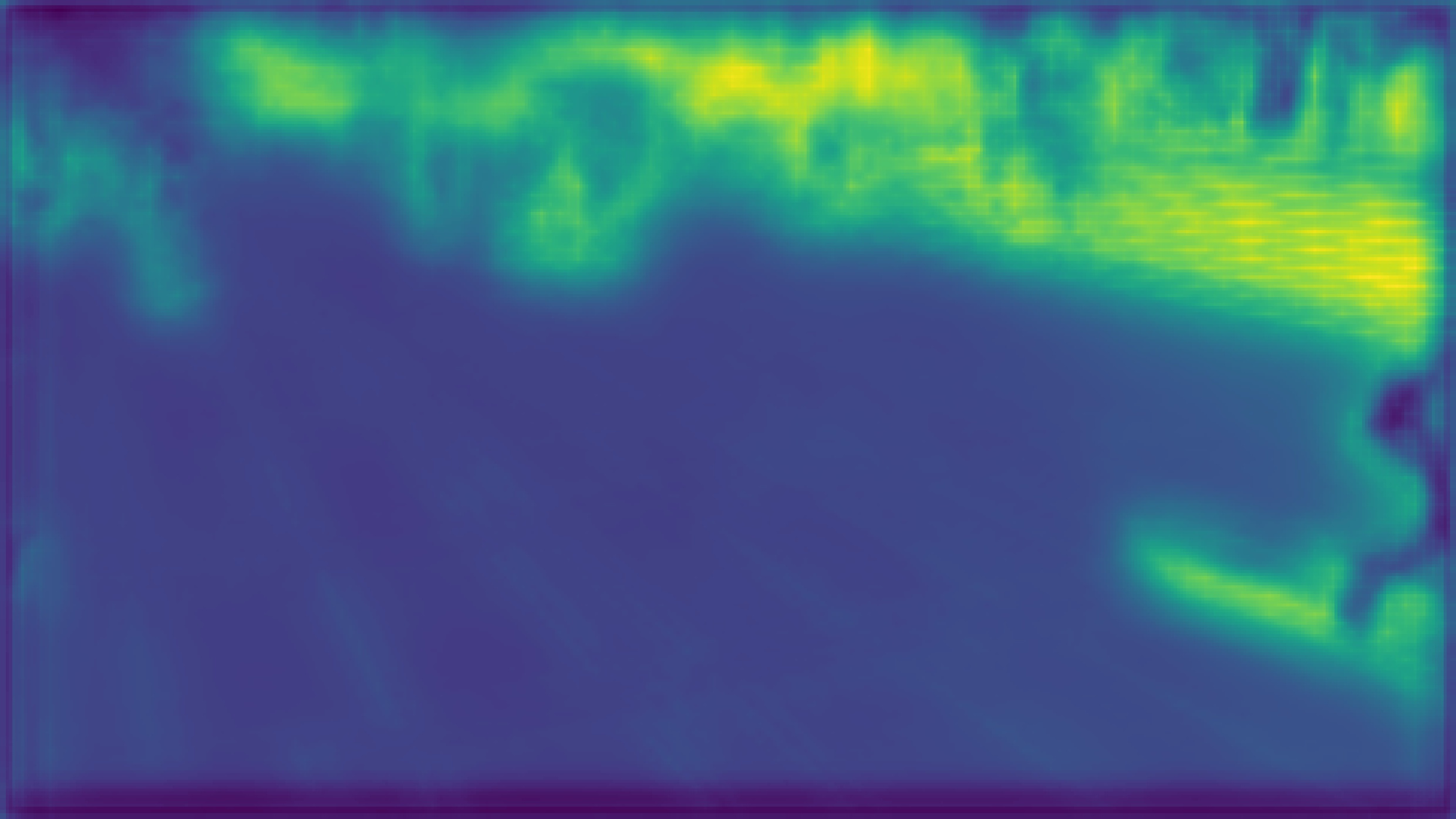} \\
				(c) & (d)\\
			\end{tabular}
			
			\begin{tabular}{ccc}
				\includegraphics[width=0.25\textwidth]{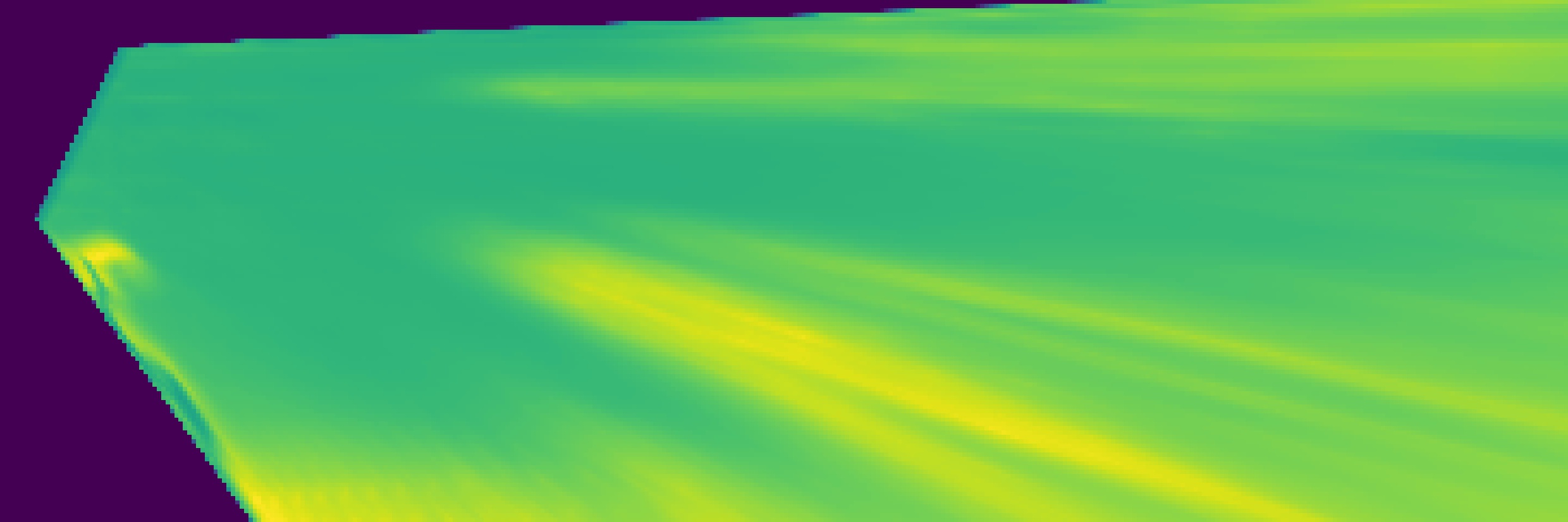} &
				\includegraphics[width=0.25\textwidth]{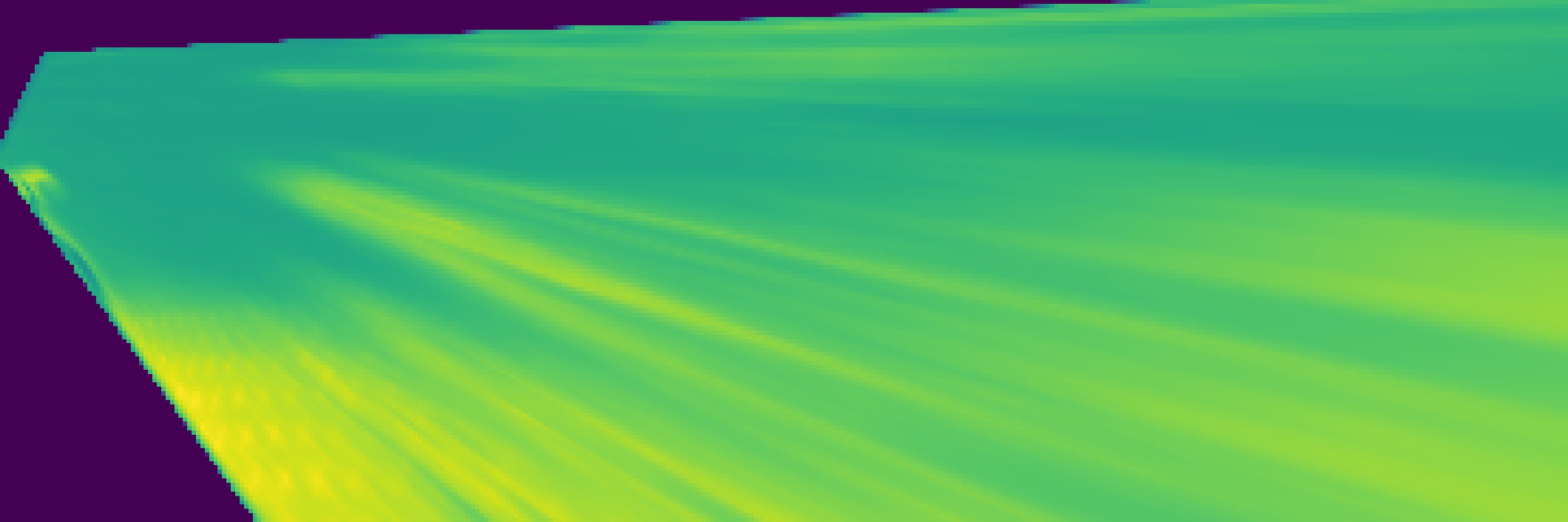} &
				\includegraphics[width=0.25\textwidth]{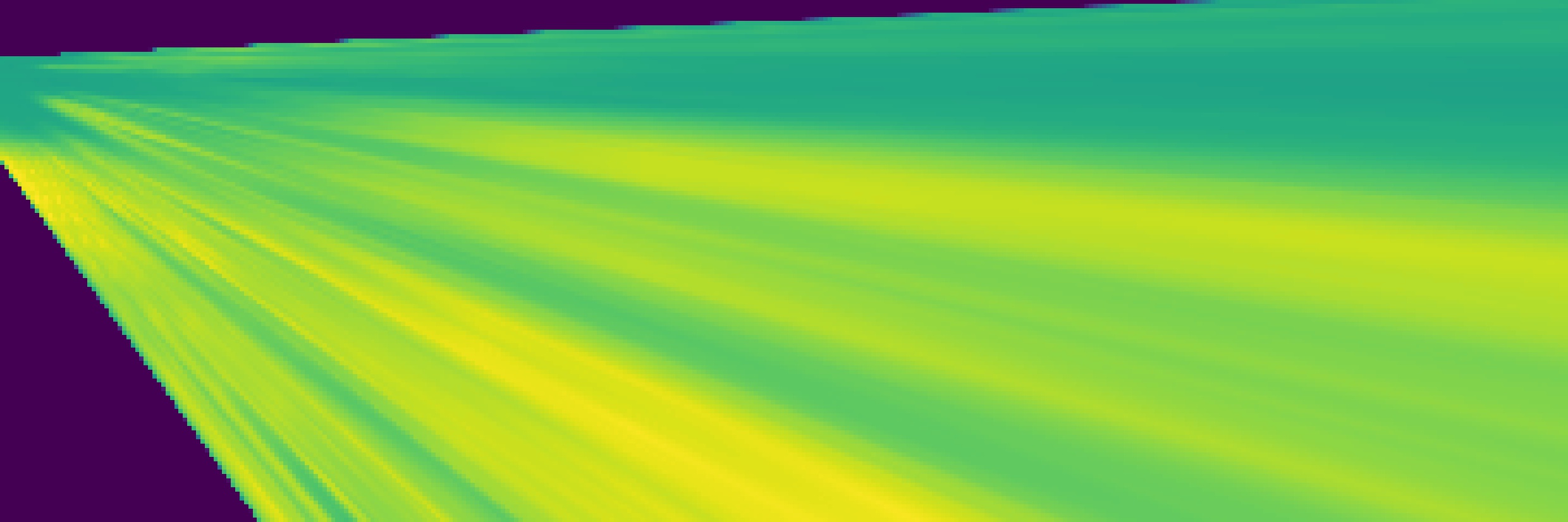} \\
				(e)&(f)&(g)\\
			\end{tabular}
			\caption{The multi-layer projections: (a) from a single camera view, (b) from two camera views, and an example of the WILDTRACK dataset, where (c) is the original image of a camera view, (d) is its feature map, and (e)-(g) are the projected feature maps at the heights of 0cm, 90cm and 180cm, respectively.
			}
			\label{f4}
		\end{center}
	\end{figure}

	The multi-layer feature projection is illustrated in Fig.~\ref{f4}(a) and (b). The projected features (or silhouette) of a pedestrian is like the shadow of that pedestrian. When the features of a pedestrian are projected from multiple views to a specific plane, they intersect at the body parts of that pedestrian at the height of that plane. By using multi-plane feature projection, the features across the height of each pedestrian can be utilized.
	An example of the multi-layer feature projection is shown in Fig.~\ref{f4}(c)-(g).
	
	The projected feature maps and the ground-plane coordinate map are concatenated for the inference of pedestrian locations. The concatenated feature maps are denoted as:
	\begin{equation}
		F = \{F_h^{c,t}, c \in[1,C], h\in\{h_1, h_2,\dots, h_M \} \}.
	\end{equation}
	where $M$ is the number of the parallel planes used for feature map projection.
	
	Since the feature maps are projected to the top view with geometric deformation, a layer of DCNv2 \cite{zhu2019deformable} is added to the top view CNN to handle the geometric deformation in the projected feature maps. The DCNv2 layer is a complementary component used with the multi-layer projection in 3DROM.

	\subsection{Loss Function}
	The loss function is the same with that of the MVDet \cite{hou2020multiview}. The network output is an occupancy probability map $\tilde{g}$. A Gaussian kernel $f(\cdot)$ is used to blur the ground-truth  pedestrian occupancy map $g$. The loss of the top view $L_t$ is the Euclidean distance between them:
	\begin{equation}
		L_t = \| \tilde{g} - f(g) \|_2.
	\end{equation}
	
	The loss function of the single view detection in camera view $c$ is:
	\begin{equation}
		L_{single}^c = \| \tilde{s}_{head}^c - f(s_{head}^c) \|_2 + \| \tilde{s}_{foot}^c - f(s_{foot}^c) \|_2,
	\end{equation}
	where $\tilde{s}_{head}^c$ and $\tilde{s}_{foot}^c$ are the single-view likelihood maps for heads and feet, respectively; $s_{head}^c$ and $s_{foot}^c$ are the ground-truth location maps for heads and feet, respectively.
	
	The overall loss for training 3DROM combines the single view loss $L_{single}$ and the top view loss $L_t$. It can be written as:
	\begin{equation}
		L_{overall} = L_t + \frac{1}{C}\sum^{C}_{c=1}{L_{single}^c}.
	\end{equation}

	\section{Experimental Results}
	\subsection{Experiment Setup}
	The proposed method has been evaluated on the EPFL WILDTRACK~\cite{WILDTRACK_dataset}\cite{chavdarova2018wildtrack}, MultiviewX~\cite{MultiviewX_dataset}\cite{hou2020multiview} and EPFL Terrace datasets~\cite{Terrace_dataset}. These three public video datasets have been widely used to evaluate multi-view pedestrian detection algorithms. Tab.~\ref{tab:dataset} shows the detailed information of these datasets.
	\begin{table}
		\begin{center}
			\caption{Datasets used for performance evaluation.}
			\label{tab:dataset}
			\scalebox{0.9}{
				\begin{tabular}{lccccccc}
					\hline
					Dataset    & \begin{tabular}[c]{@{}l@{}}Input \\ Resolusion\\ \end{tabular} &\begin{tabular}[c]{@{}l@{}}Feature \\ Resolusion\\ \end{tabular}& \begin{tabular}[c]{@{}l@{}}Training\\ Frames\end{tabular} & \begin{tabular}[c]{@{}l@{}}Testing\\ Frames\end{tabular} & \begin{tabular}[c]{@{}l@{}}AOI\\ ($m\times m$)\end{tabular} & \begin{tabular}[c]{@{}l@{}}Top View\\ Grid Size\end{tabular}&\begin{tabular}[c]{@{}l@{}}Number of 3D\\ Occlusions\end{tabular}
					\\
					\hline
					WILDTRACK  &  $1920\times1080$ & $270\times480$   & 360    & 40   & $12\times36$   & $120\times360$ &25\\
					MultiviewX & $1920\times1080$ & $270\times480$   & 360    & 40   & $16\times25$   & $160\times250$ &25\\
					Terrace    & $360\times288$  & $360\times288$   & 300    & 200  & $5.3\times5$   & $220\times150$ &20\\
					\hline
				\end{tabular}
			}
		\end{center}		
	\end{table}

	The proposed 3DROM method is based on the MVDet framework. Therefore, most of the network parameters were set to the same values as those in MVDet. ResNet-18 was used as the backbone network without using a pre-trained model. The kernel of DCNv2 used in location regression in the top view was set to a size of $2\times2$. The setup of the input image size, the feature map size, the top view grid size and the number of 3D random occlusions for each dataset are shown in Tab.~\ref{tab:dataset}. The 3D random occlusions were added to each frame before the images were input to the backbone in the training.

	For the training and testing on all the three datasets, the number of projection layers was set to $M=5$. The feature maps were projected to 5 parallel planes at the heights of 0 cm, 15 cm, 30 cm, 60 cm and 90 cm, respectively. The batch size was set to 1. The occlusion probability $p$ was set to 100\%. All experiments were carried out using one RTX-3090 GPU.

	\subsection{Qualitative Performance Evaluation}
\begin{figure}
		\centering
		\includegraphics[width=4.8in]{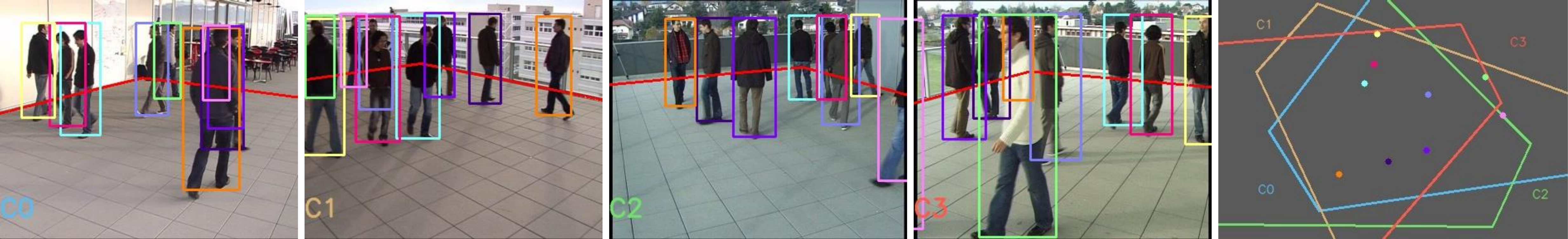}
		\caption{The detection results at frame 3225 of the EPFL Terrace dataset: from left to right, camera views C0, C1, C2, C3 and the top view. Each detected pedestrian is represented by a distinguished colour consistent across different views. The red rectangle on the ground is the AOI. The field of view of each camera is shown in the top view.
		}
		\label{f5}
	\end{figure}

	The performance of 3DROM on three datasets is demonstrated in the qualitative evaluation. Fig.~\ref{f5} shows the detection results at frame 3225 of the EPFL Terrace dataset with four camera views. The red rectangle on the ground shows the AOI region. The pedestrians outside the AOI were ignored in the detection and evaluation. The camera positions labelled in the top view are approximate ones. The colour points in the top view represent the detected pedestrians. Meanwhile, the colour of each point in the top view is consistent with the colour of the bounding boxes of the same pedestrian in all the camera views. As can be seen in Fig.~\ref{f5}, the pedestrian in the pink bounding box is completely occluded in C0, partially occluded in C1 and C2, and out of the field of view in C3. The 3DROM method can still infer the location of this pedestrian using limited pedestrian features, which demonstrates its strong detection capability in heavy occlusion.

\begin{figure}[t]
		\centering
		\includegraphics[width=4.8in]{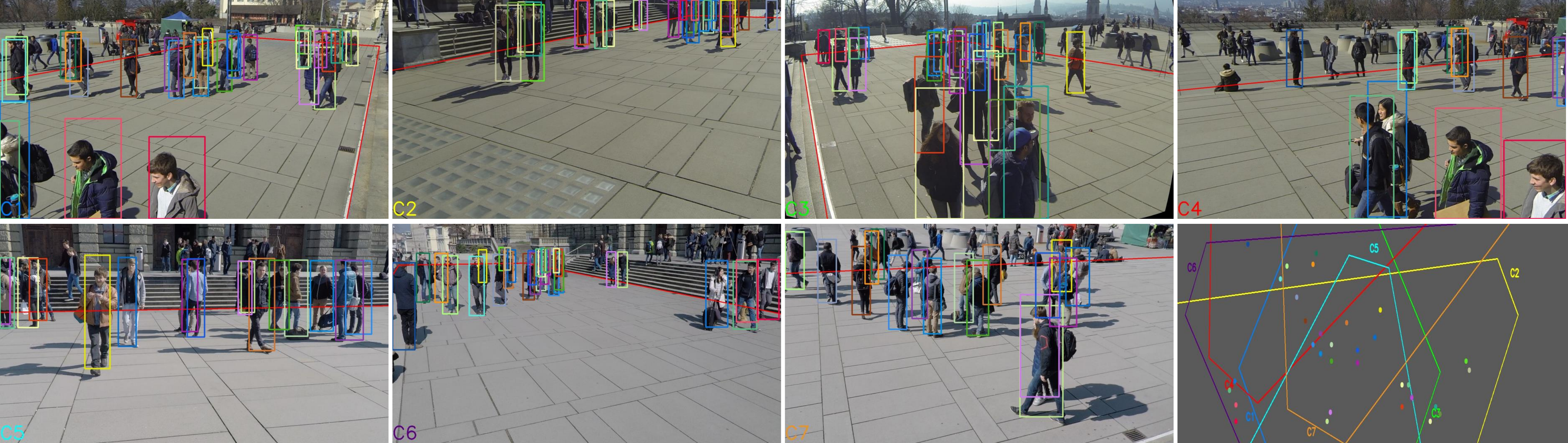}
		\caption{The detection results at frame 1960 of the EPFL WILDTRACK dataset: (top row) from left to right, camera views C1, C2, C3 and C4; (bottom row) camera views C5, C6, C7 and the top view.}
		\label{f6}
	\end{figure}

	Fig.~\ref{f6} shows the detection results at frame 1960 of the EPFL WILDTRACK dataset with seven camera views. The pedestrians stand in a group at the centre of the square and are occluded by each other. The 3DROM algorithm combines the feature information in the multi-view and multi-layer feature projections. These pedestrians are detected correctly by 3DROM.
	
\begin{figure}[t]
		\centering
		\includegraphics[width=4.8in]{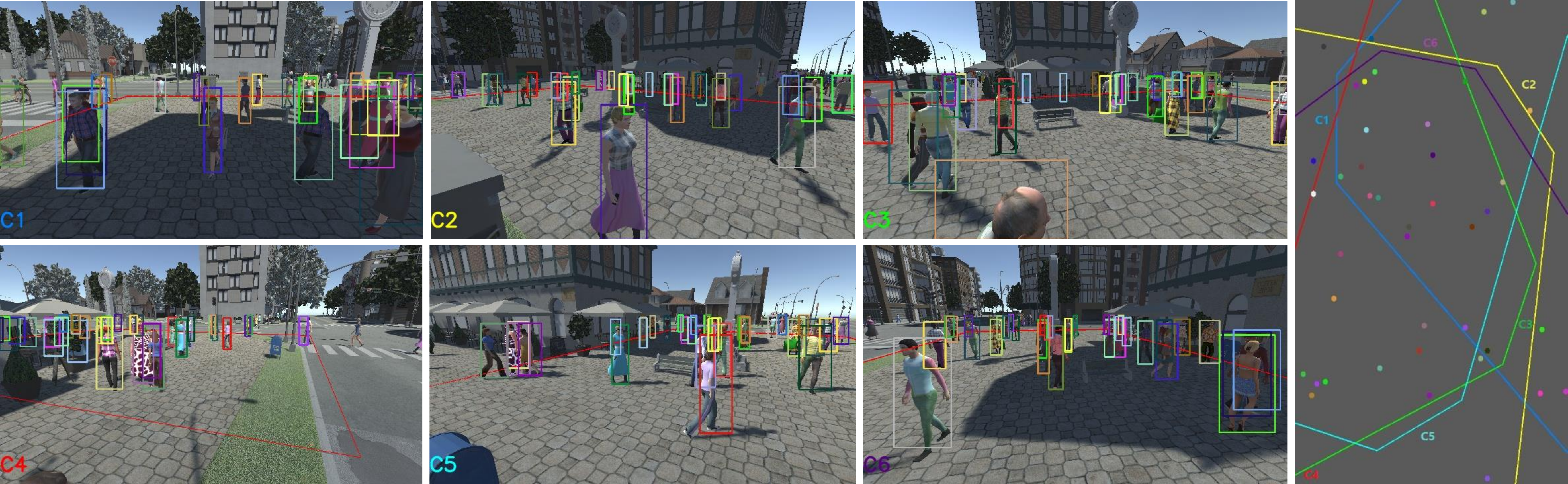}
		\caption{The detection results at frame 399 on the MultiviewX dataset: (top row) from left to right, camera views C1, C2 and C3; (bottom row) camera views C4, C5, C6 and the top view.
		}
		\label{f7}
	\end{figure}

	Fig.~\ref{f7} shows the detection results at frame 399 of the MultiviewX dataset with six camera views. A large number of pedestrians are standing very close to the border of the AOI in the top view with limited feature information. The use of multi-layer feature projection and 3D Random Occlusion in the training allows the 3DROM algorithm to detect such pedestrians accurately.
	
	\subsection{Quantitative Evaluation}
	The proposed method was evaluated using performance metrics Multiple Object Detection Accuracy (MODA) \cite{kasturi2008framework}, Multiple Object Detection Precision (MODP)~\cite{kasturi2008framework}, Precision (Prec.) and Recall, which are widely used for multi-view pedestrian detection.
	The Hungarian algorithm was used to match the detected pedestrians and ground-truth pedestrians. A distance threshold \textit{r} = 0.5m on the ground plane was used in the matching.
	
	The proposed method was compared with several state-of-the-art deep-learning based methods such as RCNN-2D/3D \cite{xu2016multi}, POM-CNN \cite{fleuret2007multicamera}, DeepMCD~\cite{chavdarova2017deep}, Deep Occlusion \cite{baque2017deep}, MVDet~\cite{hou2020multiview} and SHOT \cite{song2021stacked}, as shown in Tab.~\ref{tab:performance} in which "Eval.' indicates who made the evaluation. The MODA of the 3DROM method is increased to 93.5\%, 95.0\%, and 94.8\% in the evaluation on the WILDTRACK, MultiviewX, and Terrace datasets, respectively. Compared with the baseline algorithm MVDet that uses single-layer projections, the 3DROM increases the MODA by 5.3\%, 11.1\%, and 7.6\%, respectively. Compared with the algorithm SHOT that partly uses multi-layer projections, the 3DROM increases the MODA by 3.3\%, 6.7\% and 7.7\%, respectively. Meanwhile, the 3DROM achieves the best performance in almost all the four performance metrics.
	\begin{table}[t]
		\begin{center}
			\caption{Performance comparisons of deep multiview pedestrian detection.}
			\label{tab:performance}
			\scalebox{0.9}{
				\begin{tabular}{llllll}
					\hline
					\multicolumn{6}{c}{MultiviewX Dataset}                                    \\ \hline
					Methods             &Eval.   & MODA & MODP & Prec. & Recall \\ \hline
					RCNN-2D/3D~\cite{xu2016multi} &\cite{hou2020multiview}  & 0.187 & 0.464 & 0.635      & 0.439   \\
					DeepMCD~\cite{chavdarova2017deep}         &\cite{hou2020multiview}    & 0.700 & 0.730 & 0.857      & 0.833   \\
					Deep Occlusion~\cite{baque2017deep}   &\cite{hou2020multiview}     & 0.752 & 0.547 & 0.978      & 0.802   \\
					MVDet~\cite{hou2020multiview}           &\cite{hou2020multiview}     & 0.839 & 0.796 & 0.968      & 0.867   \\
					SHOT~\cite{song2021stacked}           &\cite{song2021stacked}     & 0.883 & 0.820 & 0.966      & 0.915   \\
					3DROM  & ours   & \textbf{0.950} & \textbf{0.849} & \textbf{0.990}      & \textbf{0.961}   \\ \hline
					\multicolumn{6}{c}{EPFL WILDTRACK Dataset}                                     \\ \hline
					RCNN-2D/3D~\cite{xu2016multi} &\cite{baque2017deep}   & 0.113 & 0.184 & 0.68       & 0.43    \\
					POM-CNN~\cite{baque2017deep}          &\cite{baque2017deep}     & 0.232 & 0.305 & 0.75       & 0.55    \\
					DeepMCD~\cite{chavdarova2017deep}           &\cite{baque2017deep}    & 0.678 & 0.642 & 0.85       & 0.82    \\
					Deep Occlusion~\cite{baque2017deep}   &\cite{baque2017deep}     & 0.741 & 0.538 & 0.95       & 0.80    \\
					MVDet~\cite{hou2020multiview}           &\cite{hou2020multiview}      & 0.882 & 0.757 & 0.947      & 0.936   \\
					SHOT~\cite{song2021stacked}           &\cite{song2021stacked}     & 0.902 & \textbf{0.765} & 0.961      & 0.940   \\
					
					3DROM   &ours    & \textbf{0.935} & 0.759 & \textbf{0.972}      & \textbf{0.962}   \\ \hline
					\multicolumn{6}{c}{EPFL Terrace Dataset}                                       \\ \hline
					RCNN-2D/3D~\cite{xu2016multi} &\cite{baque2017deep}    & -0.11 & 0.28  & 0.39       & 0.50    \\
					POM-CNN~\cite{baque2017deep}          &\cite{baque2017deep}      & 0.58  & 0.46  & 0.80       & 0.78    \\
					Deep Occlusion~\cite{baque2017deep}   &\cite{baque2017deep}      & 0.71  & 0.48  & 0.88       & 0.82    \\
					MVDet~\cite{hou2020multiview}           & ours      & 0.872 & 0.700 & 0.982      & 0.888   \\
					SHOT~\cite{song2021stacked}             &ours &0.871 &0.703	&0.989	&0.881 \\
					3DROM    & ours   & \textbf{0.948} & \textbf{0.705} & \textbf{0.997}      & \textbf{0.951}   \\ \hline
				\end{tabular}
			}
		\end{center}	
	\end{table}
	
	\textbf{Ablation Study}. In order to evaluate the contributions of each component in our model, an ablation study was carried out. The results are shown in Tab.~\ref{tab:ablation}, in which M denotes the multi-layer projection and R represents 3D Random Occlusion. As seen from the result, whichever component is added on the baseline MVDet, the performance can have a significant boost in all three datasets. When both components are used in 3DROM, the models are driven to find more robust features across multiple views, and multi-layer projection can provide sufficient features. These two components do not conflict but work better together.
	
	\begin{table}[t]
		\begin{center}
			\caption{Ablation study of 3DROM.}
			\label{tab:ablation}
			\scalebox{0.85}{
				\begin{tabular}{l|llll|llll|llll}
					\hline
					& \multicolumn{4}{c|}{MultiviewX Dataset} &  \multicolumn{4}{c|}{WILDTRACK Dataset} &  \multicolumn{4}{c}{Terrace Dataset}                                 \\ \hline
					Methods            & MODA & MODP & Prec. & Recall & MODA & MODP & Prec. & Recall & MODA & MODP & Prec. & Recall \\ \hline
					MVDet              & 0.839 & 0.796 & 0.968      & 0.867  & 0.882 & 0.757 & 0.947      & 0.936 & 0.872 & 0.700 & 0.982      & 0.888  \\
					MVDet$+$M          & 0.900 & 0.837 & 0.975      & 0.924 & 0.912 & \textbf{0.769} & 0.959      & 0.953 & 0.894 & 0.689 & 0.983      & 0.911  \\
					MVDet$+$R          & 0.898 & 0.830 & 0.986      & 0.912 & 0.923 & 0.768 & 0.964      & 0.959 & 0.915 & \textbf{0.709} & 0.994 & 0.920 \\
					3DROM              & \textbf{0.950} & \textbf{0.849} & \textbf{0.990}      & \textbf{0.961} & \textbf{0.935} & 0.759 & \textbf{0.972}  & \textbf{0.962} & \textbf{0.948} & 0.705 & \textbf{0.997}      & \textbf{0.951} \\ \hline
				\end{tabular}
			}
		\end{center}
	\end{table}
	
	\textbf{Choice of Projection Layers}. To illustrate the benefits of using five-layer feature projections, a validation study was carried out on the Terrace dataset. As reported in Tab.~\ref{tab:validation_projection}, when more than one layer is used in the feature map projection, MODA increases with the number of the projection layers.
	The experiments show that the feature projection, by using the planes below the waist height (100 cm), leads to better results than that using the planes equidistantly selected between 0 cm and the average pedestrian height 180 cm. This can be interpreted as follows: as can be seen in Fig.~\ref{f4}(a) and (b), in comparison with the ground-plane projection, the feature projection of a pedestrian on a higher plane tends to move towards the underlying camera in the top view, which projects the features, for the pedestrians who are outside of the AOI, into the AOI of the top view. Therefore, by using a projection plane at the pedestrians' heights, the features, extracted from the distant pedestrians, disturb the pedestrian detection within the AOI. The use of the projection planes below the waist is a good trade-off between the benefits of using multiple planes and the side effects.
	\begin{table}[htbp]
		\centering
		\caption{Validation of the number of projection layers (with 3D Random Occlusion applied).}
		\label{tab:validation_projection}
		\scalebox{1.0}{
			\begin{tabular}{c|l|llll}
				\hline
				Layers & Heights $(cm)$ & MODA & MODP & Prec. & Recall \\ \hline
				1 & 0   & 0.915 & 0.709 & 0.994      & 0.920   \\ \hline
				
				\multirow{2}{*}{2} & 0, 180  & 0.867 & 0.688 & 0.972     & 0.893   \\
				& 0, 60   & 0.934 & 0.708 & 0.996     & 0.937   \\ \hline
				
				\multirow{2}{*}{3} & 0, 90, 180 & 0.892 & 0.700 & 0.973     & 0.918   \\
				& 0, 60, 90  & 0.936 & 0.710 & \textbf{0.997}      & 0.938   \\ \hline
				
				\multirow{2}{*}{4} & 0, 60, 120, 180   & 0.902 & 0.697 & 0.988     & 0.912   \\
				& 0, 30, 60, 90   & 0.943  & \textbf{0.712}  & 0.996   & 0.946    \\ \hline
				\multirow{2}{*}{5} & 0, 45, 90, 135, 180  & 0.901 & 0.682 & 0.966     & 0.934   \\
				& 0, 15, 30, 60, 90   & \textbf{0.948} & 0.705  & \textbf{0.997} & \textbf{0.951}    \\ \hline
			\end{tabular}
		}
	\end{table}

	\begin{table}[b]
		\begin{center}
			\caption{Validation of the occlusion probability (with 5-layer projection applied).}
			\label{tab:validation_occlusion}
			\scalebox{0.9}{
				\begin{tabular}{l|llll|llll|llll}
					\hline
					& \multicolumn{4}{c|}{MultiviewX Dataset} &  \multicolumn{4}{c|}{WILDTRACK Dataset} &  \multicolumn{4}{c}{Terrace Dataset}                                 \\ \hline
					$p$            & MODA & MODP & Prec. & Recall & MODA & MODP & Prec. & Recall & MODA & MODP & Prec. & Recall \\ \hline
					0\%           & 0.900 & 0.837 & 0.975 & 0.924 & 0.882 & 0.757 & 0.947 & 0.936 & 0.894 & 0.689 & 0.983 & 0.911\\
					30\%          & 0.927 & \textbf{0.852} & \textbf{0.991} & 0.936 &0.920  &0.757  & \textbf{0.975} &0.944 & 0.924 & 0.697 & 0.982 & 0.941\\
					50\%          & 0.934  & 0.851  & 0.978 & 0.956 &0.923  &0.748  &0.961 &0.962 & 0.941 & 0.694 & 0.983 & \textbf{0.957}\\
					70\%          & 0.941 & 0.846 & 0.984 & 0.956 &0.928 & 0.742  & 0.967 & 0.960 & 0.944 & \textbf{0.706} & 0.994 & 0.949\\
					100\%         & \textbf{0.950} & 0.849 & 0.990 & \textbf{0.961} & \textbf{0.935} & \textbf{0.759} & 0.972 & \textbf{0.962} & \textbf{0.948} & 0.705 & \textbf{0.997} & 0.951\\ \hline
				\end{tabular}
			}
		\end{center}
	\end{table}
	
	\textbf{Validation of 3D Random Occlusion}. To investigate the role of the frequency to use 3D Random Occlusion, we tried different values of the occlusion probability \textit{p} for using 3D Random Occlusion. As reported in Tab.~\ref{tab:validation_occlusion}, MODA increases with \textit{p} and reaches the maximum value when $p = 100\%$ in all three datasets. We further compared the 3D Random Occlusion with the related Random Erasing method which was applied to each camera view independently. In this experiment, the optimal settings of Random Erasing~\cite{zhong2020random} proposed by the authors were used. In Tab.~\ref{tab:validation_augmentation}, The MODA decreases after 3D Random Occlusion is replaced by Random Erasing in all three datasets. This experiment shows the 3D Random Occlusion method can simulate the effect of Random Erasing in 3D space and is specifically designed for multi-view detection.
	
\begin{figure}
		\centering
		\includegraphics[width=3.1in]{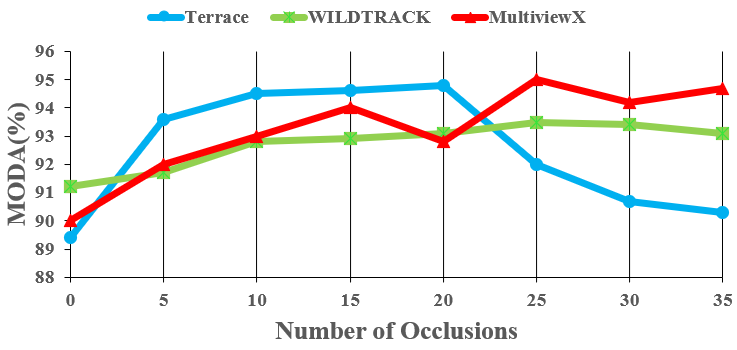}
		\caption{Parameter validation on the number of occlusions.}
		\label{f8}
	\end{figure}

	Fig.~\ref{f8} shows the validation of the number of 3D random occlusions. When occlusions are too few, the risk of overfitting increases in the training. On the other hand, too many occlusions will cover most pedestrians so that the network cannot learn effective features well. The most appropriate number of occlusions used in training correlates with the average number of pedestrians per frame and the density of pedestrians. Since the WILDTRACK and MultiviewX datasets contain more pedestrians than the Terrace, this number is greater.
	
	\begin{table}[t]
		\begin{center}
			\caption{A comparison of data augmentation schemes (with 5-layer projection applied).}
			\label{tab:validation_augmentation}
			\scalebox{0.8}{
				\begin{tabular}{l|llll|llll|llll}
					\hline
					& \multicolumn{4}{c|}{MultiviewX Dataset} &  \multicolumn{4}{c|}{WILDTRACK Dataset} &  \multicolumn{4}{c}{Terrace Dataset}                                 \\ \hline
					Methods  & MODA & MODP & Prec. & Recall & MODA & MODP & Prec. & Recall & MODA & MODP & Prec. & Recall\\ \hline
					w/o Augmentation   & 0.900 & 0.837 & 0.975 & 0.924 & 0.882 & 0.757 & 0.947 & 0.936 & 0.894 & 0.689 & 0.983 & 0.911   \\
					Random Erasing     & 0.927 & 0.847 & 0.983 & 0.943 & 0.920 & \textbf{0.766} & 0.953 & \textbf{0.967} & 0.923 & 0.692 & 0.980 & 0.943   \\
					3DROM & \textbf{0.950} & \textbf{0.849} & \textbf{0.990} & \textbf{0.961} & \textbf{0.935} & 0.759 & \textbf{0.972} & 0.962 & \textbf{0.948} & \textbf{0.705} & \textbf{0.997} & \textbf{0.951}   \\ \hline
				\end{tabular}
			}
		\end{center}
	\end{table}
	
	\section{Conclusions and Future Work}
	
	In this paper, we have proposed 3DROM for deep multiview pedestrian detection, which is based on the MVDet framework. 3D Random Occlusion provides extra training samples to the multi-view pedestrian detection network to improve the robustness in occlusion and prevent overfitting. In addition, by learning the multi-layer feature information, 3DROM can fully utilize the limited feature information from each camera view and improve pedestrian detection performance. The greatly improved performance of the 3DROM has been demonstrated in comparison with state-of-the-art methods. Future work is to find a more efficient way to fuse large-scale features and improve the across-dataset generalizability in deep-learning based multi-view pedestrian detection.
	
	\subsubsection{Acknowledgments}
	This work was supported by National Natural Science Foundation of China (NSFC) under Grant 60975082 and Xi'an Jiaotong-Liverpool University under Grant RDF-17-01-33, RDF-19-01-21 and FOSA2106045.

	\clearpage
	%
	%
	\bibliographystyle{splncs04}
	\bibliography{3DROM}
\end{document}